\documentclass[
 preprint,
 superscriptaddress,
 amsmath,amssymb,
 aps,
 floatfix
]{revtex4-2}
\usepackage{array}[=2016-10-06]
\usepackage{graphicx}
\usepackage{dcolumn}
\usepackage{bm}
\usepackage{hyperref}
\usepackage{multirow}
\usepackage{booktabs}
\usepackage{subcaption} 
\usepackage{wrapfig}
\usepackage{needspace}
\usepackage{tabularx}
\usepackage{dblfloatfix}

\begin{document}

\title{How Architecture and Training Affect TPC Representations Across Experiments}

\author{Tyler Wheeler}
\thanks{Author to whom any correspondence should be addressed.}
\email{wheeltyl@gvsu.edu}
\affiliation{Department of Computational Math, Science, and Engineering, Michigan State University, East Lansing, MI, USA}
\affiliation{Department of Physics, Grand Valley State University, Allendale, Michigan, USA}

\author{Michelle P. Kuchera}
\affiliation{Department of Physics, Davidson College, Davidson, NC, USA}
\affiliation{Department of Mathematics and Computer Science, Davidson College, Davidson, NC, USA}

\author{Raghuram Ramanujan}
\affiliation{Department of Mathematics and Computer Science, Davidson College, Davidson, NC, USA}

\author{William Sieland}
\affiliation{Department of Physics, Davidson College, Davidson, NC, USA}

\author{Ryan Krupp}
\affiliation{Department of Computational Math, Science, and Engineering, Michigan State University, East Lansing, MI, USA}

\author{Yassid Ayyad}
\affiliation{Department of Particle Physics, Universidad de Santiago de Compostela, Santiago de Compostella, Spain}
\affiliation{Facility for Rare Isotope Beams, Michigan State University, East Lansing, Michigan 48824, USA}

\author{Daniel Bazin}
\affiliation{Facility for Rare Isotope Beams, Michigan State University, East Lansing, Michigan 48824, USA}
\affiliation{Department of Physics and Astronomy, Michigan State University, East Lansing, Michigan 48824, USA}

\author{Connor L.~Cross}
\affiliation{Department of Mathematics and Computer Science, Davidson College, Davidson, NC, USA}

\author{Hoi Yan Ian Heung}
\affiliation{Department of Physics, Davidson College, Davidson, NC, USA}

\author{Andrew J.~Jones}
\affiliation{Department of Mathematics and Computer Science, Davidson College, Davidson, NC, USA}

\author{Ruchi Mahajan}
\affiliation{Department of Physics and Astronomy, University of Kentucky, Lexington, KY, USA}

\author{Saiprasad Ravishankar}
\affiliation{Department of Computational Math, Science, and Engineering, Michigan State University, East Lansing, MI, USA}

\author{Pranjal Singh}
\affiliation{Facility for Rare Isotope Beams, Michigan State University, East Lansing, Michigan 48824, USA}
\affiliation{Department of Physics and Astronomy, Michigan State University, East Lansing, Michigan 48824, USA}

\author{Benjamin Votaw}
\affiliation{Department of Physics, Davidson College, Davidson, NC, USA}

\author{Chris Wrede}
\affiliation{Facility for Rare Isotope Beams, Michigan State University, East Lansing, Michigan 48824, USA}
\affiliation{Department of Physics and Astronomy, Michigan State University, East Lansing, Michigan 48824, USA}


\begin{abstract}

In recent years, deep-learning efforts have increasingly shifted towards foundation model approaches. In experimental physics, this allows for models and learned representations to be reused beyond the experiments in which they were originally developed. This work evaluates the reusability of model representations across experiments and detector systems via probes on frozen encoders. These frozen probes reveal task-relevant structure available before downstream
adaptation, providing a complementary diagnostic to fine-tuning. Together with
random-weight controls, they distinguish contributions between architecture
and encoder training that downstream performance alone cannot resolve.

Time projection chamber (TPC) data provide a useful testbed for studying this problem because events from different TPC systems can be represented as variable-length sparse tensors, while the underlying detector geometries, event topologies, and scientific tasks can differ substantially. We investigate whether fixed-dimensional TPC event representations can be reused across classification tasks, experiments, and detector systems. Sparse ResNet and PointNet-style encoders produce 512-dimensional embeddings for four experimental datasets from the GADGET~II TPC and the Active-Target TPC. Randomly initialized encoders isolate the contribution from the  architecture before supervised encoder training. We then train each encoder on a simple classification task, freeze its parameters, and train a new linear or nonlinear probe on the latent representation for each downstream task. We find that this architecture-induced structure remains useful across experiments and detector systems. The randomly initialized PointNet-style representation is already highly informative on several tasks. The two architectures organize their embedding spaces differently, but neither exhibits a large, systematic loss of utility in the cross-detector setting. These results show that the architecture  is a major source of task-relevant structure in TPC embeddings and should be treated explicitly when assessing representation learning and developing reusable detector models.



\end{abstract}

\keywords{time projection chambers, representation learning, transfer learning, foundation models, sparse convolutional neural networks, nuclear physics}

\maketitle

\section{Introduction}

Time projection chambers (TPCs) are a common detector type for high resolution spatial reconstruction of charged-particle paths through a gaseous or liquid medium. This work focuses on gas-filled low-energy nuclear physics TPCs. As charged reaction products traverse a gas-filled volume, they ionize the medium and liberate electrons that drift under a uniform electric field toward a segmented readout plane, where the drift time and pad position can be used to reconstruct the complete three-dimensional particle trajectory. Further, the charge collected along the trajectory recovers the energy loss. This combination of full-track reconstruction and particle identification allows event selection and analysis to happen offline, rather than relying on a real-time hardware trigger. This makes TPCs particularly well-suited for studies of exotic and rare nuclei where reaction rates are low. As the use of TPCs has grown more widespread, their raw output---sparse, variable-length point clouds of spatial coordinates and ionization charge---has become one of the richest and increasingly common data products available in the field.

Machine learning (ML) methods have emerged as one of the standard components of the toolkit for data analysis across nuclear and particle physics \citep{Boehnlein2022MLNuclear}. It is thus no surprise that they have been successfully applied to a range of tasks arising in TPC experiments such as particle classification \citep{microboone,Kuchera2019ATTPC,Robles2025PointSetLArTPC}, unsupervised event identification \cite{Solli2021ATTPC}, semantic segmentation for track identification \cite{Domine2020SparseLArTPC}, rare-event searches \cite{Dey2025ATTPCPointCloud,Wheeler2025GADGET}, background rejection \cite{NEXT}, and kinematic reconstruction \cite{Zhao2025dNdxRW}. The event signatures of interest in these studies vary with the physics task, experimental conditions, and detector geometry and the ML solutions have consequently been developed in a bespoke fashion for individual analyses.

To avoid this cost of repeatedly training models from scratch for each new application or domain, the field of ML has recently focused its attention on \emph{foundation models} \cite{foundationmodels} --- generalist models that learn effective representations of their vast training data in a self-supervised manner that can then be specialized for a range of downstream tasks with minimal amounts of labeled data. Recent representation learning studies in physics have shown that such pretrained models can support new tasks or domains, including masked particle modeling and OmniLearn for collider data \citep{Golling2024MaskedParticle,Mikuni2025OmniLearn}, PoLAr-MAE and Panda for LArTPC data \citep{Young2025PoLArMAE,Young2025Panda}, and transfer between collider-detector geometries \citep{Mokhtar2025CrossDetector}. This work uses deep learning models trained on data gathered from two different TPC systems (the Active-Target \cite{BRADT201765} and the GADGET II \cite{PhysRevC.110.035807} TPCs) to explore the degree to which representations learned in one setting transfer effectively to another. Specifically, we investigate the benefits of representation reuse when the learning task changes within an experiment, when the experiment changes within a detector, and when the detector itself changes. In each case, we explore whether a fixed-length embedding of the variable-length, permutation invariant input point cloud retains information from which a new task-specific classifier can be learned. This approach is immediately relevant to TPC analyses, where labeled samples may be limited and event selection often precedes computationally intensive reconstruction or fitting. A preliminary version of this study appeared at the 2025 Machine Learning and the Physical Sciences workshop at NeurIPS \cite{Wheeler2025SparseTPC}.

\section{Background}
\subsection{Transfer learning and representation reuse}

Transfer learning uses information acquired from a source task or domain to support learning in a target task or domain \citep{Pan2010Transfer}.  In deep
networks, this information is commonly transferred either by fine-tuning the pretrained network on target data, or by freezing the encoder and training a new task-specific model on its outputs \citep{Yosinski2014Transferable}.  The
latter approach tests the reuse of the learned representation without changing the encoder, which we focus on in this work.

\subsection{Representation capacity and architecture effects}
Model capacity broadly describes the range and complexity of functions that a
model can represent.  It is influenced not only by the size of the network, but also the architecture choice. The architecture imposes an inductive bias by determining how information can be combined. We consider two broad families of architectures in this work, both of which are well-suited to point cloud inputs: sparse convolutional neural networks (CNNs) and Pointnet-style models \cite{Qi2017PointNet}. Both are implemented using the Minkowski Engine library \cite{Choy_2019_CVPR} which introduces efficient sparse tensor operators. This allows us to port traditional, high-performing CNN architectures originally developed for computer vision applications to TPC data, as well as extend the capability of traditional Pointnet to natively handle variable-length events. We study encoders developed with both architectures to better understand whether they organize event information differently.


For a frozen encoder, probe performance provides an operational measure of the
task-relevant information accessible in its embedding. A linear probe tests whether information is linearly separable, while a nonlinear probe can exploit more complex structure in the same fixed representation.  Randomly initialized
networks act as nonlinear feature maps, analogous to random
feature methods \citep{Rahimi2007RandomFeatures}.  Comparing probes trained on
random and pretrained encoder outputs therefore helps separate structure induced by the architecture from the additional organization introduced by supervised
pretraining.

\begin{figure}[htbp]
\centering

\begin{subfigure}[b]{0.49\textwidth}
    \centering
    \includegraphics[width=\textwidth]{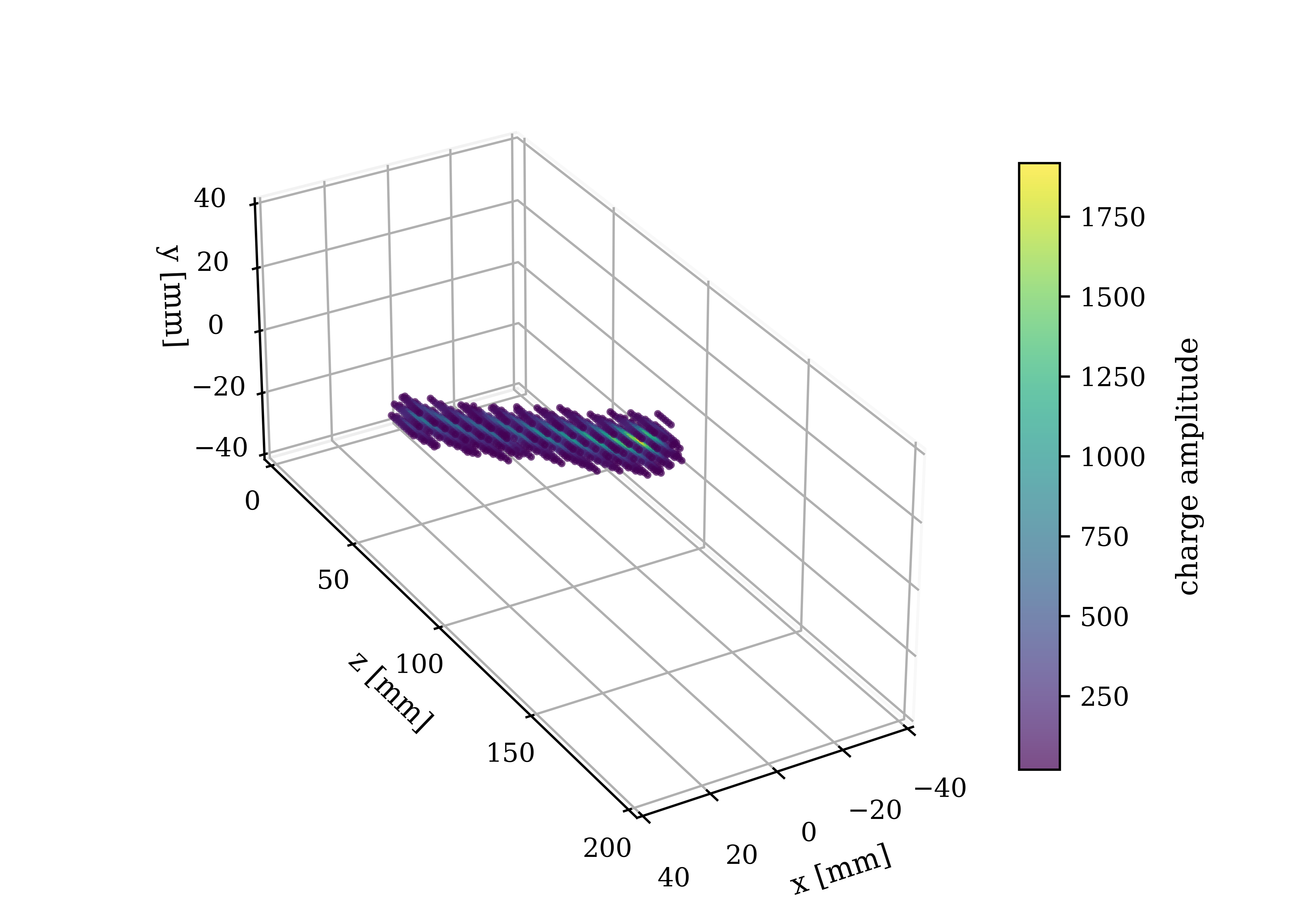}
    \caption{GADGET II Event}
    \label{fig:gadget_event}
\end{subfigure}
\hfill
\begin{subfigure}[b]{0.40\textwidth}
    \centering
    \includegraphics[width=\textwidth]{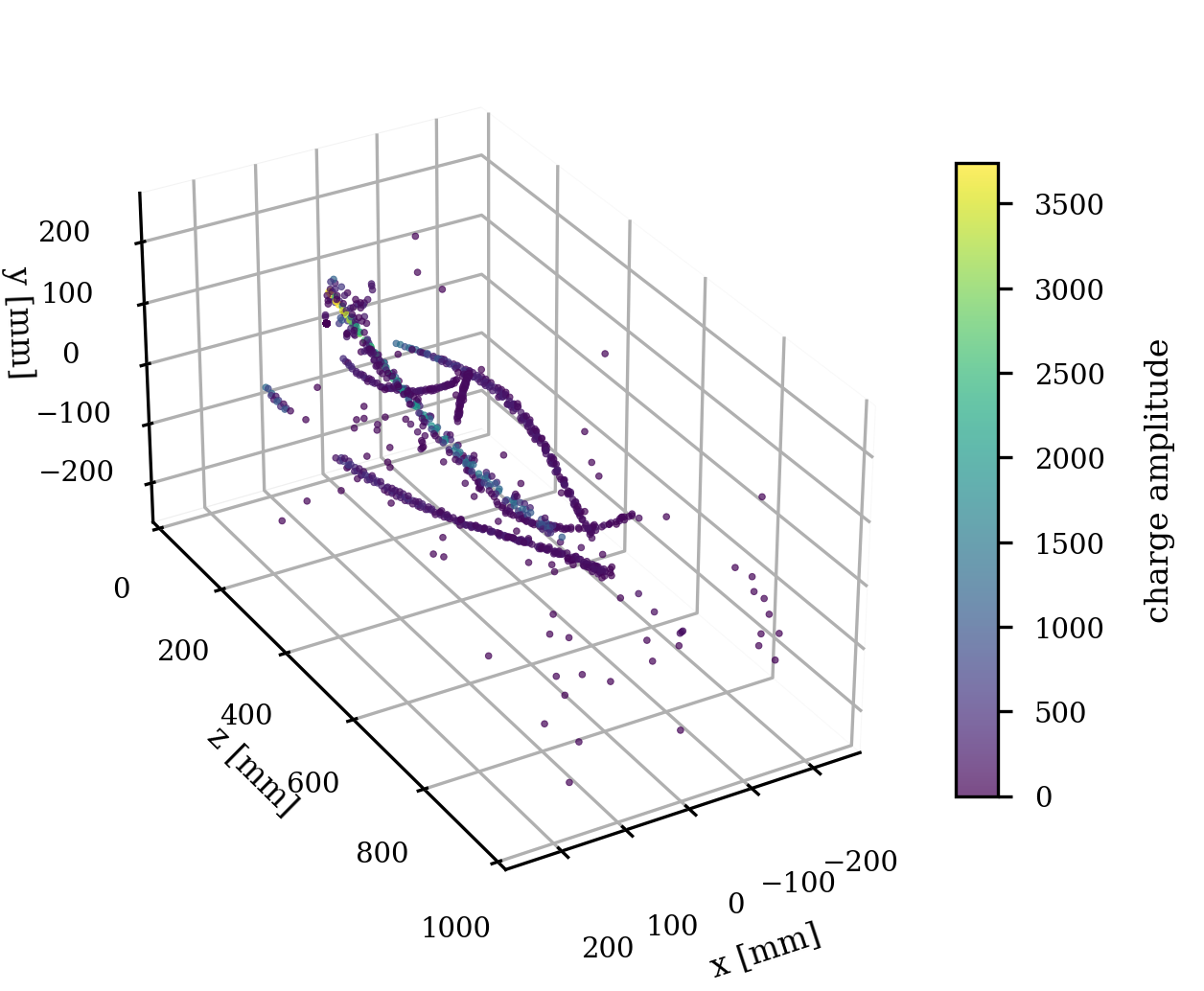}
    \caption{AT-TPC Event}
    \label{fig:attpc_event}
\end{subfigure}

\caption{Example event topologies from the two TPC systems used in this work. (a) A \textsuperscript{20}Mg, 1600 keV beta-delayed proton event from the GADGET II TPC. (b) A 4-track event from the AT-TPC.}
\label{fig:side_by_side}

\end{figure}

\subsection{TPCs in this study}
We consider data from two TPCs with substantially different detector geometries and experimental use cases: the GAseous Detector with GErmanium Tagging (GADGET~II) TPC \citep{PhysRevC.110.035807}, 
and the Active-Target TPC (AT-TPC) \citep{BRADT201765,AYYAD2020161341}.
Detector details are given below, and example event topologies from the two systems are shown in Fig.~\ref{fig:side_by_side}. Particular attention should be given to the contrast between the event topologies, as they appear to be from unrelated domains. This difference makes the observed transfer of learned representations between the two detectors a nontrivial result, and motivates the use of sparse tensor methods as a detector-agnostic approach to TPC data. 
A summary of key detector parameters is given in Table \ref{tab:tpc_comparison}.

\subsubsection{GADGET II TPC}
The GADGET~II TPC is a radioactive ion beam implant-decay detector optimized for low-energy $\beta$-delayed charged-particle spectroscopy. Radioactive ions are thermalized in the detector gas and decay within the active volume. Charged particles emitted in the decay ionize the gas along their trajectories, and the resulting ionization electrons drift in a uniform electric field toward a position-sensitive resistive-anode MICROMEGAS readout plane. The measured pad position, drift time, and collected charge provide the information needed to reconstruct the three-dimensional event topology.

The detector has a cylindrical active volume with a 40 cm drift region and an active pad-plane diameter of approximately 8 cm. It is typically operated with P10 gas at 800 torr, with a nominal drift field of 150 V/cm and an amplification field of approximately 30 kV/cm across the MICROMEGAS amplification gap. Under these operating conditions, the measured electron drift velocity is $5.44 \pm 0.03$ cm/$\mu$s. The TPC is also coupled with DEGAi, an array of high-purity germanium detectors used for $\gamma$-ray spectroscopy and to verify the implantation location of the beam. A schematic of the GADGET~II detection system is shown in Fig.\ref{fig:gadget}.

The readout plane is segmented into 1024 pads, consisting of 1016 central measurement pads and 8 surrounding veto pads. The measurement pads are $2.2 \times 2.2$ mm$^2$, and the veto pads are used to identify events in which charged particles leave the active volume and deposit only partial energy. Signals from the MICROMEGAS pads are digitized with the GET electronics system, which provides one electronic channel per pad. For each triggered event, the fired pads and their associated charge traces form a naturally sparse detector response.

\subsubsection{AT-TPC}
The AT-TPC is a large active-target time projection chamber designed for nuclear reaction studies in inverse kinematics. In active-target operation, the detector gas serves simultaneously as the reaction target and the tracking medium. As the incoming beam slows in the gas, it can react with the target nuclei, and both the beam trajectory and charged reaction products are tracked through their ionization. This enables measurements with low-intensity radioactive beams while maintaining high effective target thickness.

\begin{figure}[t]
\centering

\begin{subfigure}[t]{0.48\linewidth}
\centering
\includegraphics[width=\linewidth]{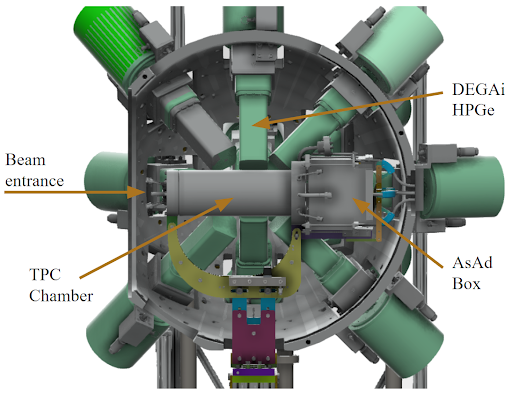}
\caption{GADGET II}
\label{fig:gadget}
\end{subfigure}
\hfill
\begin{subfigure}[t]{0.48\linewidth}
\centering
\includegraphics[width=\linewidth]{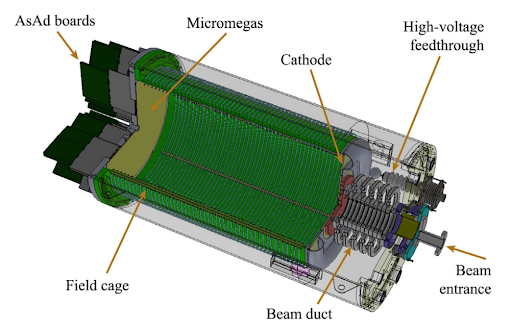}
\caption{AT-TPC}
\label{fig:attpc}
\end{subfigure}

\caption{(a) Schematic of the GAseous Detector with GErmanium Tagging (GADGET) II TPC—a detector optimized for measuring low-energy $\beta$-delayed particle decays, surrounded by a germanium array. (b) Schematic of the Active-Target TPC (AT-TPC)—a detector designed for nuclear reaction studies in inverse kinematics operated in a solenoid magnet providing up to 4 T of magnetic field.}
\label{fig}
\end{figure}

The AT-TPC has a cylindrical geometry with a length  of approximately 100 cm and a diameter of approximately 58 cm. The beam is injected along the detector axis from the cathode end cap. Ionization electrons drift toward a highly segmented pad plane, where the drift time provides the coordinate along the beam direction and the pad location provides the transverse coordinates. The detector can be coupled with SOLARIS \citep{Kay2018}, a large solenoid magnet that provides fields up to 4 Tesla, causing charged-particle trajectories to be curved and enabling particle-rigidity measurements and improved reconstruction of charged-particle trajectories. A schematic of the AT-TPC system is shown in Fig.\ref{fig:attpc}.

Charge amplification at the pad plane is provided by a hybrid micro-pattern gaseous detector system consisting of a MICROMEGAS coupled to a multi-layer Thick Gas Electron Multiplier. The pad plane is segmented into 10,240 gold-plated triangular pads. It contains an inner hexagonal region of 6,144 pads
with a height of 0.5~cm, surrounded by an outer region of 4,096 pads
with a height of 1.0~cm. The smaller inner pads provide finer spatial
resolution near the beam axis, where reaction vertices generally occur.

\begin{table}[t]
\centering
\small
\setlength{\tabcolsep}{6pt}
\renewcommand{\arraystretch}{1.15}

\caption{Comparison of the two TPC systems used in this work.}
\label{tab:tpc_comparison}

\begin{tabular}{
>{\raggedright\arraybackslash}p{0.27\linewidth}
>{\raggedright\arraybackslash}p{0.31\linewidth}
>{\raggedright\arraybackslash}p{0.31\linewidth}
}
\toprule
\textbf{Feature} &
\textbf{GADGET~II TPC} &
\textbf{AT-TPC} \\
\midrule

Primary use case &
$\beta$-delayed charged-particle decays &
Inverse-kinematics reactions \\

Detector mode &
Implant-decay TPC &
Active-target TPC \\

Active/drift length &
40~cm drift region &
100~cm drift region \\

Active diameter &
8~cm pad-plane diameter &
55~cm pad-plane diameter \\

Gas/target role &
Stopping/decay medium &
Reaction target and tracking medium \\

Typical gas &
P10 at 800~Torr &
Experiment-dependent target gas (e.g., $^{4}$He, $^{2}$H) \\

Electric field &
150~V/cm drift field &
100~V/cm drift field \\

Magnetic field &
None &
Up to 4 T solenoid \\

Charge amplification &
Resistive-anode MICROMEGAS &
MICROMEGAS + MTHGEM \\

Pad plane &
1,016 square pads &
10,240 triangular pads \\

Pad size &
$2.2 \times 2.2$~mm$^2$ &
0.5~cm inner / 1.0~cm outer pad heights \\

\bottomrule
\end{tabular}
\end{table}

\section{Methods}
\subsection{Data and Learning Tasks}

We use data from four experiments spanning two TPC systems at the Facility for Rare Isotope Beams (FRIB): two implant-decay datasets collected with the GADGET II TPC and two active-target datasets collected with the AT-TPC. For both detector systems, events are represented as sparse $(x,y,z,q)$ points, with the pad-plane positions providing the transverse coordinates and the electron drift time providing the longitudinal coordinate. However, the charge feature, $q$, corresponds to the integrated charge on each pad for GADGET, and charge amplitude for the AT-TPC. For each detector, a comparatively simple classification task is used to train the encoder, after which the learned representations are evaluated using more complex downstream tasks. These evaluations are organized into three groups: within-experiment transfer, in which the experiment remains the same but the classification task changes; cross-experiment transfer, in which the experiment changes while the detector remains the same; and cross-detector transfer, in which both the experiment and detector system change. The datasets and learning tasks are described below, and Table~\ref{tab:data_tasks} provides a summary of class compositions and event counts.

\subsubsection{GADGET II $^{20}$Mg dataset}\mbox{}\par
\noindent FRIB experiment E21072 used a $\sim$60 MeV/u $^{20}$Mg beam delivered to the GADGET~II TPC. An upstream aluminum degrader was used to tune the beam energy such that the ions stopped near the center of the detector, which was filled with P10 gas at $\sim$800 Torr. The beam was delivered in alternating 110~ms beam-on and 110~ms beam-off periods, with $\beta$-delayed charged-particle emissions measured during the beam-off window. 

Two learning tasks were defined for the $^{20}$Mg dataset. The first is a binary particle-identification task distinguishing proton and alpha-particle events. This comparatively simple task is used to train the $^{20}$Mg encoders. The resulting representations are then probed in three settings: within-experiment on a more complex three-class $^{20}$Mg task, cross-experiment on the $^{21}$Mg task, and cross-detector on the $^{16}$O and $^{10}$B tasks.

The three-class $^{20}$Mg task distinguishes 770~keV protons, 1596~keV protons, and 2153~keV alpha particles. Because this task uses the same isotope and detector as the binary training task while introducing a more fine-grained classification problem, it serves as the within-experiment downstream probe for the $^{20}$Mg-trained models.

Labels for both tasks were assigned using gates on range-versus-energy distributions. Individual proton and alpha-particle decay branches form localized populations in this space, allowing particle species and individual decay energies to be selected. For the binary task, one gate was placed around the proton population and another around the alpha-particle population using approximately one hour of data. For the three-class task, separate tight gates were placed around the three selected decay branches. 

\subsubsection{GADGET II $^{21}$Mg dataset}\mbox{}\par
\noindent FRIB experiment E25058 used a $\sim$60 MeV/u $^{21}$Mg beam delivered to the GADGET~II TPC. An upstream aluminum degrader was used to tune the beam energy such that the ions stopped near the center of the detector, which was filled with P10 gas at $\sim$800 Torr. The beam was delivered in alternating 250~ms beam-on and 250~ms beam-off periods, with $\beta$-delayed charged-particle emissions measured during the beam-off window. 

A three-class classification task was defined for the $^{21}$Mg dataset, distinguishing 1285~keV protons, 1930~keV protons, and 2153~keV alpha particles. This task serves as a cross-experiment probe for encoders trained on the $^{20}$Mg binary task and as a cross-detector probe for encoders trained on the $^{16}$O binary task. Labels were assigned using tight gates around the corresponding populations in range-versus-energy space. 

\subsubsection{AT-TPC \(^{16}\mathrm{O}+\alpha\) dataset}\mbox{}\par
\noindent FRIB experiment E20020 used a $^{16}$O beam at 10~MeV/u incident on a $^4$He gas active target at 700~Torr. The AT-TPC was operated inside the SOLARIS solenoid magnet with a nearly uniform magnetic field of 3~T \citep{RegueiraCastro2025}. 

Two track-multiplicity classification tasks were defined for the $^{16}$O dataset. In the three-class task, events were grouped into 0--2, 3, and 4--5 track classes based on categories relevant to the experimental analysis. A simpler binary task was constructed by combining these categories into 0--2-track and 3--5-track classes.

The binary task is used to train the $^{16}$O encoders. The learned representations are then evaluated within-experiment on the three-class $^{16}$O task, cross-experiment on the $^{10}$B task, and cross-detector on the three-class $^{20}$Mg and $^{21}$Mg tasks. Labels were assigned through visual inspection of the detector events, with the number of tracks manually identified for each event.

\subsubsection{AT-TPC \(^{10}\mathrm{B}+d\) dataset}\mbox{}\par
\noindent FRIB experiment E20009 used the AT-TPC with a deuterium gas active target. Events were visually inspected and manually labeled according to track multiplicity. The classification task used in this work distinguishes three-track from four-track events, with the selected categories motivated by the event topologies relevant to the experimental analysis.

The $^{10}$B task is used only as a downstream probe. It provides a cross-experiment test for representations learned from the $^{16}$O binary task and a cross-detector test for representations learned from the $^{20}$Mg binary task. 

\begin{table*}[!tb]
\centering
\small
\setlength{\tabcolsep}{4pt}
\renewcommand{\arraystretch}{1.15}

\begin{tabularx}{\textwidth}{
@{}
l
l
c
>{\raggedright\arraybackslash}X
>{\raggedright\arraybackslash}X
c
r
@{}
}
\toprule
\textbf{Dataset} &
\textbf{Detector} &
\textbf{Isotope} &
\textbf{Task name} &
\textbf{Classes} &
\shortstack{\textbf{Events per}\\\textbf{class}} &
\shortstack{\textbf{Total}\\\textbf{events}} \\
\midrule
E21072 &
GADGET~II &
$^{20}$Mg &
$^{20}$Mg, 2 class &
Proton; alpha &
25,078; 9,501 &
34,579 \\

E21072 &
GADGET~II &
$^{20}$Mg &
$^{20}$Mg, 3 class &
770~keV p; 1596~keV p; 2153~keV $\alpha$ &
18,770; 6,308; 9,501 &
34,579 \\

\addlinespace

E25058 &
GADGET~II &
$^{21}$Mg &
$^{21}$Mg, 3 class &
1285~keV p; 1930~keV p; 2153~keV $\alpha$ &
1,261; 1,351; 3,363 &
5,975 \\

\midrule

E20020 &
AT-TPC &
$^{16}$O &
$^{16}$O$+\alpha$, 2 class &
0--2 tracks; 3--5 tracks &
1,184; 1,274 &
2,458 \\

E20020 &
AT-TPC &
$^{16}$O &
$^{16}$O$+\alpha$, 3 class &
0--2 tracks; 3 tracks; 4--5 tracks &
1,184; 476; 798 &
2,458 \\

\addlinespace

E20009 &
AT-TPC &
$^{10}$B &
$^{10}$B$+d$, 2 class &
3 tracks; 4 tracks &
2,310; 2,151 &
4,461 \\

\bottomrule
\end{tabularx}
\caption{Summary of the datasets used in this work.}
\label{tab:data_tasks}
\end{table*}

\subsection{Network Architectures and Training}


We investigated frozen representations produced by sparse ResNet14 and PointNet-style encoders implemented with the Minkowski Engine. Each event is represented by a variable-length set of three-dimensional spatial coordinates and associated ionization charge. In the final models, the normalized spatial coordinates and log-scaled charge, $(x,y,z,q)$, are supplied as four feature channels; the spatial coordinates also specify the locations of the sparse inputs. This representation therefore retains both the event topology and the measured charge information without padding events to a common length.

Both encoders produce 512-dimensional event representations but differ substantially in capacity. The frozen ResNet14 encoder contains 21,423,616 trainable parameters, whereas the frozen PointNet-style encoder contains 83,840. Including the source-task classification heads, the complete models contain 21,424,642 and 348,034 trainable parameters, respectively.

\subsubsection{ResNet backbone}

\begin{figure*}[!tb]
    \centering

    \includegraphics[
        width=\textwidth
    ]{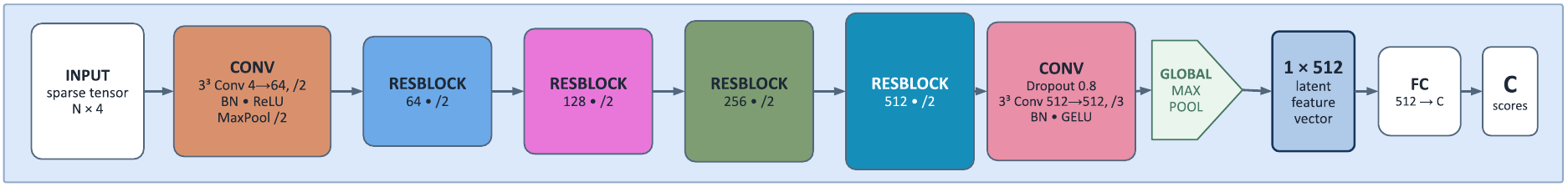}

    \vspace{0.8em}

    \includegraphics[
        width=0.72\textwidth
    ]{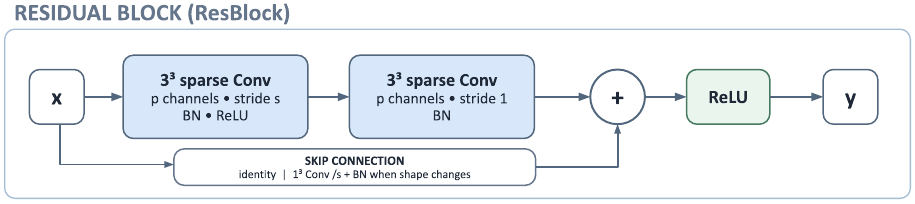}

    \caption{
    Sparse ResNet14 architecture used to produce event-level TPC
    embeddings. The lower panel shows the structure of each residual
    block.
    }
    \label{fig:resnet_arch}
\end{figure*}

We use a convolutional neural network, specifically a shallow residual network (ResNet14) \citep{he2015deepresiduallearningimage}, adapted for sparse inputs. Its main components are summarized in Figure~\ref{fig:resnet_arch}. Convolutions operate in the three spatial dimensions, while $q$ is treated as a feature channel. The design includes an initial stem, four residual stages, and a final pre-pooling block followed by global max pooling and a fully connected head. This configuration uses ten convolutional layers and one linear layer. We found that deeper variants (e.g., ResNet50 with bottleneck blocks) increased training time without clear performance gains.


\begin{figure*}[!tb]
    \centering
    \includegraphics[width=\textwidth]{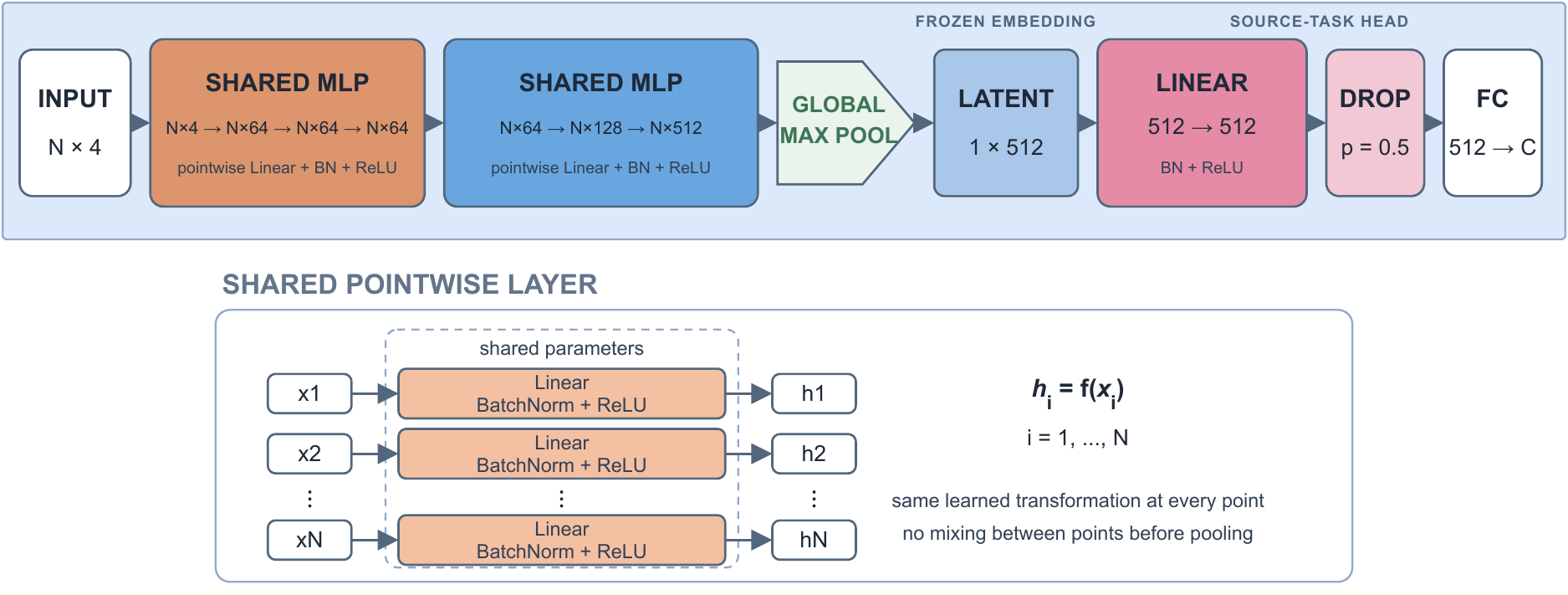}
    \caption{Minkowski PointNet-style architecture used to embed TPC events. Top: a shared multilayer perceptron is applied independently to each of the input points. Global max pooling aggregates the resulting $N\times512$ features into the 512-dimensional event embedding. The post-pooling linear layer, dropout, and final classifier are used during source-task training but are not included in the extracted embedding. Bottom: each pointwise layer applies the same learned transformation independently to every active input; information is combined across inputs only by global pooling.}
    \label{fig:pointnet_arch}
\end{figure*}
\subsubsection{PointNet-style backbone}
An alternative backbone is comprised of a Minkowski-Engine adaptation of the PointNet architecture \cite{pointnet,Choy_2019_CVPR}. The main components are outlined in Figure~\ref{fig:pointnet_arch}. The hybrid approach follows the PointNet approach of shared point-wise transformations, accomplishing this in a sparse tensor framework through $1\times1$ convolutions followed by symmetric global pooling for permutation invariance. The sparse tensor implementation extends the capabilities of the original PointNet implementation to allow for variable length inputs, which enables the preservation of the variable length structure of our detector events.

\subsubsection{Training and optimization}
For each encoder training task, five models were
trained using five-fold cross-validation. In each instance, one
fold was held out for evaluation of the trained encoder, while the remaining
data were divided into training and validation subsets. Spatial coordinate and charge scaling parameters were determined using only the corresponding training and validation data. 
Weighted random sampling was applied to the training subset
for both architectures to reduce the effect of class imbalance.

Training was monitored using validation loss, and the checkpoint with the
lowest validation loss was retained for each fold. Complete optimization,
regularization, early-stopping, and augmentation settings are reported in
Appendix~\ref{app:training}.

\subsection{Latent Embedding Analysis}

Each trained or randomly initialized encoder was frozen, and its
512-dimensional event embeddings were used as inputs to two support-vector
machine (SVM) probes. The linear probe used a linear SVM with regularization
parameter \(C=1\). The nonlinear probe used a radial-basis-function (RBF)
kernel with \(C=1\) and the scikit-learn \texttt{gamma=scale} convention.
These hyperparameters were fixed for every architecture, encoder-training task, and downstream dataset; they were not optimized separately for individual evaluations.

Each embedding dimension was standardized using statistics from the training set, with the same transformation applied to the test set. For within-experiment evaluations, the probe was trained using the combined training and validation embeddings for a given fold and evaluated on
that fold's held-out test embeddings. For cross-experiment and cross-detector evaluations, the downstream dataset was divided using a stratified 80:20
training--test split.

A separate probe was trained for each encoder run and downstream task.
Accuracy and macro-averaged F1 were calculated on the held-out test
partition. Reported values are the mean and standard deviation across the
five runs. The na\"ive baseline predicted the most frequent class in the training set
and was evaluated on the same held-out test set as the SVM probes.

\section{Results}
\label{sec:results}
\subsection{Latent Probes}
The linear and RBF probes assess complementary aspects of the learned representations. The linear probe measures how readily downstream class information can be accessed using linear decision boundaries, whereas the RBF probe permits a more flexible nonlinear readout. Their relative performance therefore provides evidence about how the task-relevant class structure is organized within each embedding space.

\subsubsection{Within-Experiment Task Transfer}
Within each experiment, probes trained on the random-encoder embeddings already substantially outperform the na\"ive classifier, indicating that the architectures themselves provide useful task-relevant representations before supervised encoder training. Encoder training yields additional gains, but these are more pronounced for ResNet than for PointNet, whose random representation is already especially effective for the $^{20}\mathrm{Mg}$ task. The limited benefit from the RBF probe for $^{20}\mathrm{Mg}$, particularly with PointNet, indicates that its class structure is already largely linearly separable. The larger RBF gains for $^{16}\mathrm{O}+\alpha$, especially after training, indicate useful class structure that is not captured by a single linear decision boundary. Detailed results are reported in Table~\ref{tab:in_domain_probes}.
\begin{table*}[!tb]
\centering
\scriptsize
\setlength{\tabcolsep}{3pt}
\renewcommand{\arraystretch}{1.12}
\begin{tabular}{@{}lllcccc@{}}
\toprule
\textbf{Evaluation task} & \textbf{Encoder training} & \textbf{Architecture}
& \multicolumn{2}{c}{\textbf{Linear SVM}} & \multicolumn{2}{c}{\textbf{RBF SVM}} \\
\cmidrule(lr){4-5}\cmidrule(lr){6-7}
&&& \textbf{Accuracy} & \textbf{Macro-F1} & \textbf{Accuracy} & \textbf{Macro-F1} \\
\midrule
\multirow{5}{*}{$^{20}\mathrm{Mg}$, 3 class} & Na\"ive baseline & --- & \(0.5428 \pm 0.0000\) & \(0.2346 \pm 0.0000\) & \(0.5428 \pm 0.0000\) & \(0.2346 \pm 0.0000\) \\
 & Random weights & ResNet & \(0.9109 \pm 0.0052\) & \(0.8840 \pm 0.0068\) & \(0.9321 \pm 0.0061\) & \(0.9077 \pm 0.0091\) \\
 & Random weights & PointNet & \(0.9979 \pm 0.0005\) & \(0.9977 \pm 0.0004\) & \(0.9933 \pm 0.0013\) & \(0.9928 \pm 0.0015\) \\
 & $^{20}\mathrm{Mg}$, 2 class & ResNet & \(0.9782 \pm 0.0059\) & \(0.9732 \pm 0.0073\) & \(0.9803 \pm 0.0042\) & \(0.9753 \pm 0.0053\) \\
 & $^{20}\mathrm{Mg}$, 2 class & PointNet & \(\mathbf{0.9999 \pm 0.0001}\) & \(\mathbf{0.9999 \pm 0.0001}\) & \(\mathbf{0.9999 \pm 0.0001}\) & \(\mathbf{0.9999 \pm 0.0001}\) \\
\midrule
\multirow{5}{*}{$^{16}\mathrm{O}+\alpha$, 3 class} & Na\"ive baseline & --- & \(0.4817 \pm 0.0006\) & \(0.2167 \pm 0.0002\) & \(0.4817 \pm 0.0006\) & \(0.2167 \pm 0.0002\) \\
 & Random weights & ResNet & \(0.6008 \pm 0.0081\) & \(0.5107 \pm 0.0107\) & \(0.6646 \pm 0.0121\) & \(0.4873 \pm 0.0070\) \\
 & Random weights & PointNet & \(0.8008 \pm 0.0173\) & \(0.7579 \pm 0.0197\) & \(0.8390 \pm 0.0081\) & \(0.8044 \pm 0.0101\) \\
 & $^{16}\mathrm{O}+\alpha$, 2 class & ResNet & \(0.7010 \pm 0.0266\) & \(0.6264 \pm 0.0280\) & \(0.7819 \pm 0.0262\) & \(0.6957 \pm 0.0394\) \\
 & $^{16}\mathrm{O}+\alpha$, 2 class & PointNet & \(\mathbf{0.8352 \pm 0.0142}\) & \(\mathbf{0.7941 \pm 0.0187}\) & \(\mathbf{0.8885 \pm 0.0079}\) & \(\mathbf{0.8623 \pm 0.0091}\) \\
\bottomrule
\end{tabular}
\caption{Within-experiment probe performance. Values are the mean $\pm$ standard deviation over five runs. Boldface marks the largest mean in each evaluation-task and metric column. The na\"ive baseline predicts the most frequent class in the training set.}
\label{tab:in_domain_probes}
\end{table*}

\subsubsection{Cross-Experiment Transfer}
Changing the experiment while retaining the detector does not eliminate the task-relevant structure present in the random-encoder embeddings. Across both the GADGET~II and AT-TPC transfer pairs, the random PointNet representations remain particularly informative, with encoder training providing a comparatively smaller additional benefit. ResNet benefits more substantially from encoder training on the other experiment. For the $^{10}\mathrm{B}+d$ task, the stronger trained performance with an RBF probe further suggests that some transferred class structure is accessible through nonlinear, rather than purely linear, boundaries. Detailed results are given in Table~\ref{tab:cross_experiment_probes}.
\begin{table*}[!tb]
\centering
\scriptsize
\setlength{\tabcolsep}{3pt}
\renewcommand{\arraystretch}{1.12}
\begin{tabular}{@{}lllcccc@{}}
\toprule
\textbf{Evaluation task} & \textbf{Encoder training} & \textbf{Architecture}
& \multicolumn{2}{c}{\textbf{Linear SVM}} & \multicolumn{2}{c}{\textbf{RBF SVM}} \\
\cmidrule(lr){4-5}\cmidrule(lr){6-7}
&&& \textbf{Accuracy} & \textbf{Macro-F1} & \textbf{Accuracy} & \textbf{Macro-F1} \\
\midrule
\multirow{5}{*}{$^{21}\mathrm{Mg}$, 3 class} & Na\"ive baseline & --- & \(0.5632 \pm 0.0000\) & \(0.2402 \pm 0.0000\) & \(0.5632 \pm 0.0000\) & \(0.2402 \pm 0.0000\) \\
 & Random weights & ResNet & \(0.8524 \pm 0.0051\) & \(0.7985 \pm 0.0065\) & \(0.8499 \pm 0.0142\) & \(0.7922 \pm 0.0221\) \\
 & Random weights & PointNet & \(0.9955 \pm 0.0025\) & \(0.9942 \pm 0.0031\) & \(0.9945 \pm 0.0027\) & \(0.9937 \pm 0.0030\) \\
 & $^{20}\mathrm{Mg}$, 2 class & ResNet & \(0.9506 \pm 0.0026\) & \(0.9246 \pm 0.0042\) & \(0.9622 \pm 0.0065\) & \(0.9422 \pm 0.0098\) \\
 & $^{20}\mathrm{Mg}$, 2 class & PointNet & \(\mathbf{0.9967 \pm 0.0032}\) & \(\mathbf{0.9958 \pm 0.0039}\) & \(\mathbf{0.9985 \pm 0.0006}\) & \(\mathbf{0.9983 \pm 0.0008}\) \\
\midrule
\multirow{5}{*}{$^{10}\mathrm{B}+d$, 2 class} & Na\"ive baseline & --- & \(0.5174 \pm 0.0000\) & \(0.3410 \pm 0.0000\) & \(0.5174 \pm 0.0000\) & \(0.3410 \pm 0.0000\) \\
 & Random weights & ResNet & \(0.5205 \pm 0.0244\) & \(0.5145 \pm 0.0240\) & \(0.5187 \pm 0.0072\) & \(0.4278 \pm 0.0127\) \\
 & Random weights & PointNet & \(0.8723 \pm 0.0123\) & \(0.8722 \pm 0.0122\) & \(0.8965 \pm 0.0148\) & \(0.8963 \pm 0.0148\) \\
 & $^{16}\mathrm{O}+\alpha$, 2 class & ResNet & \(0.6267 \pm 0.0261\) & \(0.6261 \pm 0.0263\) & \(0.6891 \pm 0.0123\) & \(0.6891 \pm 0.0124\) \\
 & $^{16}\mathrm{O}+\alpha$, 2 class & PointNet & \(\mathbf{0.8985 \pm 0.0089}\) & \(\mathbf{0.8985 \pm 0.0089}\) & \(\mathbf{0.9496 \pm 0.0065}\) & \(\mathbf{0.9495 \pm 0.0065}\) \\
\bottomrule
\end{tabular}
\caption{Cross-experiment probe performance. Values are the mean $\pm$ standard deviation over five runs. Boldface marks the largest mean in each evaluation-task and metric column. The na\"ive baseline predicts the most frequent class in the training set.}
\label{tab:cross_experiment_probes}
\end{table*}
\subsubsection{Cross-detector Transfer}
Cross-detector reuse follows the same architecture-dominated pattern observed in the preceding settings rather than showing a pronounced penalty from changing detector systems. Random-encoder embeddings remain strongly informative, particularly for PointNet, while supervised encoder training generally provides a further benefit despite the detector mismatch. These gains are clearest for ResNet and when nonlinear structure is accessed with the RBF probe. PointNet changes comparatively little on the GADGET~II tasks because its random representation is already near the performance ceiling. Detailed results are reported in Table~\ref{tab:cross_detector_probes}.

\begin{table*}[!tb]
\centering
\scriptsize
\setlength{\tabcolsep}{3pt}
\renewcommand{\arraystretch}{1.12}
\begin{tabular}{@{}lllcccc@{}}
\toprule
\textbf{Evaluation task} & \textbf{Encoder training} & \textbf{Architecture}
& \multicolumn{2}{c}{\textbf{Linear SVM}} & \multicolumn{2}{c}{\textbf{RBF SVM}} \\
\cmidrule(lr){4-5}\cmidrule(lr){6-7}
&&& \textbf{Accuracy} & \textbf{Macro-F1} & \textbf{Accuracy} & \textbf{Macro-F1} \\
\midrule
\multirow{5}{*}{$^{16}\mathrm{O}+\alpha$, 3 class} & Na\"ive baseline & --- & \(0.4817 \pm 0.0000\) & \(0.2167 \pm 0.0000\) & \(0.4817 \pm 0.0000\) & \(0.2167 \pm 0.0000\) \\
 & Random weights & ResNet & \(0.6008 \pm 0.0081\) & \(0.5107 \pm 0.0107\) & \(0.6646 \pm 0.0121\) & \(0.4873 \pm 0.0070\) \\
 & Random weights & PointNet & \(0.8008 \pm 0.0173\) & \(0.7579 \pm 0.0197\) & \(0.8390 \pm 0.0081\) & \(0.8044 \pm 0.0101\) \\
 & $^{20}\mathrm{Mg}$, 2 class & ResNet & \(0.6955 \pm 0.0083\) & \(0.6162 \pm 0.0060\) & \(0.7817 \pm 0.0068\) & \(0.6960 \pm 0.0172\) \\
 & $^{20}\mathrm{Mg}$, 2 class & PointNet & \(\mathbf{0.8248 \pm 0.0115}\) & \(\mathbf{0.7841 \pm 0.0175}\) & \(\mathbf{0.8720 \pm 0.0110}\) & \(\mathbf{0.8489 \pm 0.0135}\) \\
\midrule
\multirow{5}{*}{$^{10}\mathrm{B}+d$, 2 class} & Na\"ive baseline & --- & \(0.5174 \pm 0.0000\) & \(0.3410 \pm 0.0000\) & \(0.5174 \pm 0.0000\) & \(0.3410 \pm 0.0000\) \\
 & Random weights & ResNet & \(0.5205 \pm 0.0244\) & \(0.5145 \pm 0.0240\) & \(0.5187 \pm 0.0072\) & \(0.4278 \pm 0.0127\) \\
 & Random weights & PointNet & \(0.8723 \pm 0.0123\) & \(0.8722 \pm 0.0122\) & \(0.8965 \pm 0.0148\) & \(0.8963 \pm 0.0148\) \\
 & $^{20}\mathrm{Mg}$, 2 class & ResNet & \(0.6426 \pm 0.0168\) & \(0.6420 \pm 0.0167\) & \(0.6885 \pm 0.0081\) & \(0.6883 \pm 0.0080\) \\
 & $^{20}\mathrm{Mg}$, 2 class & PointNet & \(\mathbf{0.8889 \pm 0.0121}\) & \(\mathbf{0.8888 \pm 0.0121}\) & \(\mathbf{0.9355 \pm 0.0121}\) & \(\mathbf{0.9354 \pm 0.0121}\) \\
\midrule
\multirow{5}{*}{$^{20}\mathrm{Mg}$, 3 class} & Na\"ive baseline & --- & \(0.5428 \pm 0.0000\) & \(0.2346 \pm 0.0000\) & \(0.5428 \pm 0.0000\) & \(0.2346 \pm 0.0000\) \\
 & Random weights & ResNet & \(0.9109 \pm 0.0052\) & \(0.8840 \pm 0.0068\) & \(0.9321 \pm 0.0061\) & \(0.9077 \pm 0.0091\) \\
 & Random weights & PointNet & \(0.9979 \pm 0.0005\) & \(0.9977 \pm 0.0004\) & \(0.9933 \pm 0.0013\) & \(0.9928 \pm 0.0015\) \\
 & $^{16}\mathrm{O}+\alpha$, 2 class & ResNet & \(0.9451 \pm 0.0143\) & \(0.9357 \pm 0.0155\) & \(0.9842 \pm 0.0058\) & \(0.9808 \pm 0.0068\) \\
 & $^{16}\mathrm{O}+\alpha$, 2 class & PointNet & \(\mathbf{0.9994 \pm 0.0003}\) & \(\mathbf{0.9992 \pm 0.0004}\) & \(\mathbf{0.9997 \pm 0.0003}\) & \(\mathbf{0.9997 \pm 0.0003}\) \\
\midrule
\multirow{5}{*}{$^{21}\mathrm{Mg}$, 3 class} & Na\"ive baseline & --- & \(0.5632 \pm 0.0000\) & \(0.2402 \pm 0.0000\) & \(0.5632 \pm 0.0000\) & \(0.2402 \pm 0.0000\) \\
 & Random weights & ResNet & \(0.8524 \pm 0.0051\) & \(0.7985 \pm 0.0065\) & \(0.8499 \pm 0.0142\) & \(0.7922 \pm 0.0221\) \\
 & Random weights & PointNet & \(\mathbf{0.9955 \pm 0.0025}\) & \(\mathbf{0.9942 \pm 0.0031}\) & \(0.9945 \pm 0.0027\) & \(0.9937 \pm 0.0030\) \\
 & $^{16}\mathrm{O}+\alpha$, 2 class & ResNet & \(0.8790 \pm 0.0183\) & \(0.8430 \pm 0.0197\) & \(0.9374 \pm 0.0062\) & \(0.9146 \pm 0.0069\) \\
 & $^{16}\mathrm{O}+\alpha$, 2 class & PointNet & \(0.9923 \pm 0.0036\) & \(0.9901 \pm 0.0048\) & \(\mathbf{0.9982 \pm 0.0012}\) & \(\mathbf{0.9979 \pm 0.0015}\) \\
\bottomrule
\end{tabular}
\caption{Cross-detector probe performance. Values are the mean $\pm$ standard deviation over five runs. Boldface marks the largest mean in each evaluation-task and metric column. The na\"ive baseline predicts the most frequent class in the training set.}
\label{tab:cross_detector_probes}
\end{table*}

\begin{figure*}[t]
    \centering
    \includegraphics[width=\textwidth]{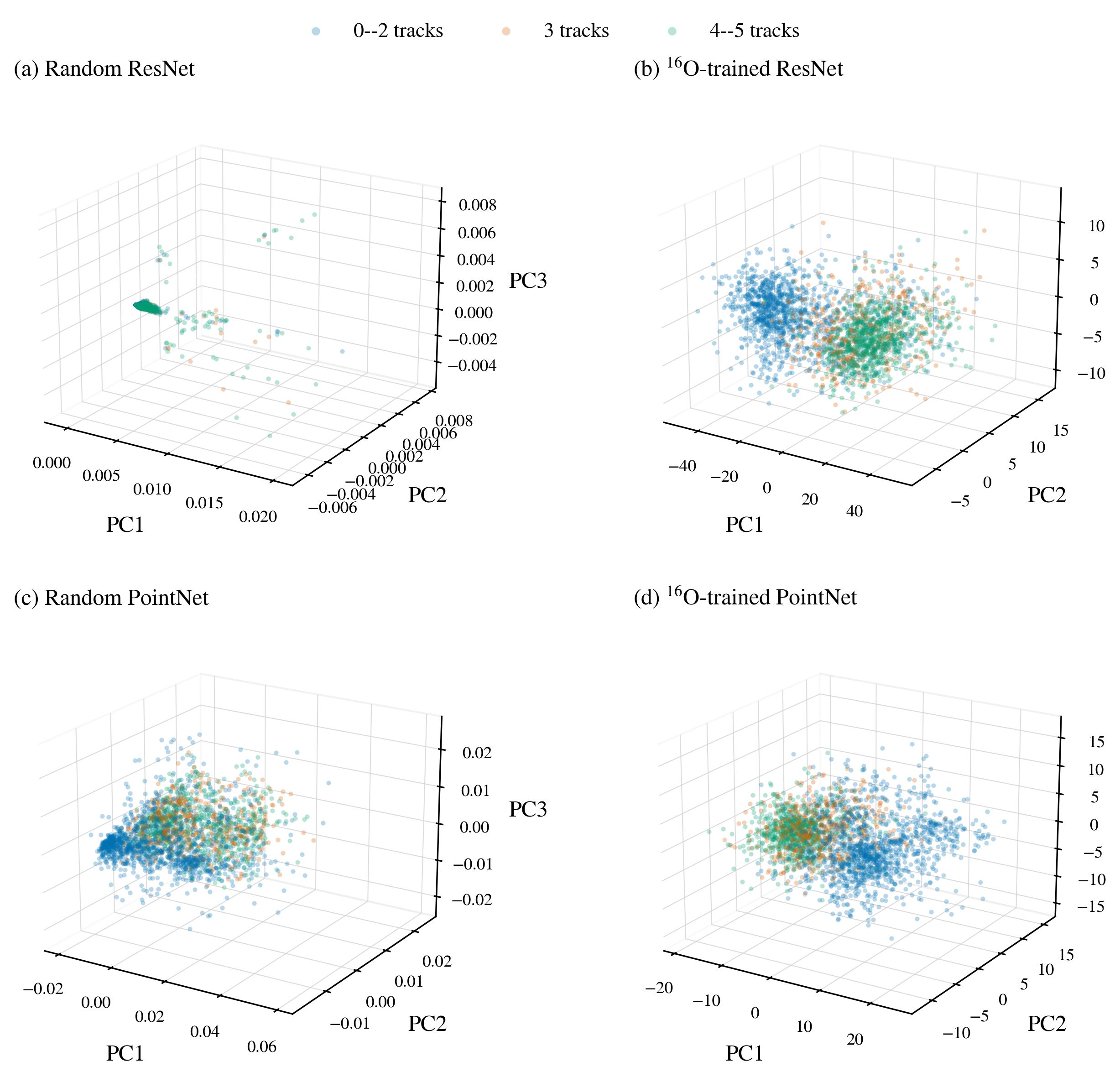}
    \caption{
    PCA projections of the latent representations for the
    $^{16}\mathrm{O}+\alpha$ three-class task. Panels compare randomly initialized and $^{16}$O-trained ResNet and PointNet-style encoders. Points are colored according to track-multiplicity class.}
    \label{fig:pca_o16}
\end{figure*}

\begin{figure*}[t]
    \centering
    \includegraphics[width=\textwidth]{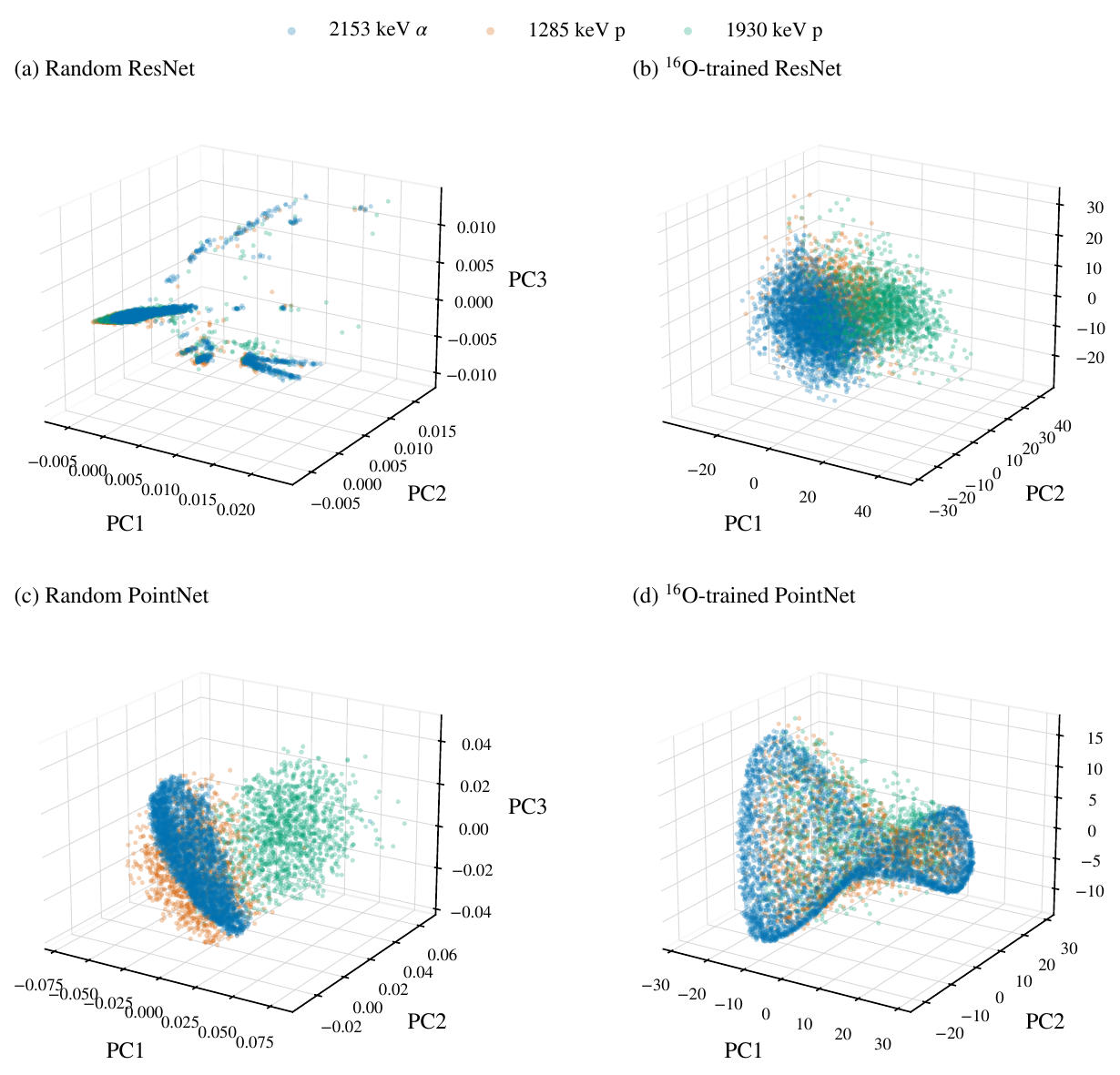}
    \caption{
    PCA projections of the latent representations for the
    $^{21}$Mg three-class task. Panels compare randomly initialized and $^{16}$O-trained ResNet and PointNet-style encoders. Points are colored according to the three decay-event classes. 
    }
    \label{fig:pca_mg21}
\end{figure*}

\subsection{PCA Visualizations}
Principal Component Analysis (PCA) provides a qualitative view of the embedding structure complementary to the probe results. Here, we show representative PCA projections for a subset of the evaluated embeddings rather than the full set of model–task combinations. Figures \ref{fig:pca_o16} and \ref{fig:pca_mg21} show the leading three principal components of the latent representations for the $^{16}\mathrm{O}+\alpha$ and $^{21}$Mg three-class tasks, respectively.

For the $^{16}\mathrm{O}+\alpha$ target task (Fig. \ref{fig:pca_o16}), the two randomly initialized encoders produce visibly different organizations of the same events. The random ResNet representation is concentrated near a compact region with a relatively small number of separated events, whereas the random PointNet representation forms a broader, more continuous distribution with visible class structure. Training on the $^{16}$O binary task substantially reorganizes both embedding spaces and makes the distinction between the 0--2-track class and the higher-multiplicity classes more apparent.

The $^{21}$Mg projections (Fig. \ref{fig:pca_mg21}) further illustrate that the two architectures encode the same downstream events with substantially different latent geometries. The random ResNet representation exhibits several narrow and separated structures, while the random PointNet representation produces broader class regions. Following $^{16}$O training, the ResNet representation becomes more diffuse, with partial separation among the three particle-energy classes. The trained PointNet representation instead occupies a broad, continuous manifold.

Together, these projections show that similar downstream probe performance does not imply similar latent-space geometry. Because PCA identifies directions of greatest variance rather than directions that maximize class separation, we interpret these visualizations qualitatively rather than as an additional measure of classification performance.

\section{Discussion}

We find that the architecture itself supplies a
large fraction of the task-relevant structure in these TPC representations.
For linear-probe macro-F1, the improvement from the modal classifier to a
probe on random-encoder embeddings is larger than the subsequent improvement
from supervised encoder training in all 16 evaluated architecture--task
settings. The same pattern holds in 14 of 16 RBF-probe comparisons.
This structure remains useful across changes in task, experiment, and
detector.  Encoder training generally adds further information, but the gain varies across architectures and tasks.  The gain is largest and most consistent for ResNet, whereas PointNet often begins from a strong random
baseline and consequently has less room to improve.

The random-encoder results demonstrate the importance of architecture choice.  However, both architectures embed informative event-level information.  

The PCA projections reinforce that the architectures embed events in a structurally different way, even when their probes attain similar performance.
These findings highlight the diagnostic value of frozen-representation
studies. 
Frozen probes complement downstream fine-tuning by measuring
task-relevant structure that is already accessible before adaptation.
Combined with architecture-matched random-weight controls, they help
distinguish the contributions of architectural bias and encoder training that
end-to-end performance combines. Reporting such diagnostics alongside
fine-tuning results can provide a more complete evaluation of reusable
scientific representations.


\begin{acknowledgments}
We thank the GADGET and AT-TPC collaborations for their efforts in collecting and providing the experimental data used in this work.
\end{acknowledgments}

\section*{Funding}
This work supported by the U.S. Department of Energy, Office of Science, under award No.DE-SC0024587. This work supported by the U.S. National Science Foundation OAC-2311263
. This material is based upon work supported by the U.S. Department of Energy, Office of Science, Office of Nuclear Physics and used resources of the Facility for Rare Isotope Beams (FRIB) Operations, which is a DOE Office of Science User Facility under Award Number DE-SC0023633.

\section*{Data Availability}
The processed data used to generate the results presented in this work will be made publicly available at the time of publication in accordance with the FRIB Data Management and Sharing Plan. Access to additional FRIB experimental data is subject to the applicable data-sharing policies and approval of the experiment co-spokespersons.

\bibliographystyle{apsrev4-2}
\bibliography{r} 

\appendix
\section{Training Details}
\label{app:training}

In each five-fold experiment, fold \(i\) was used as the held-out test fold,
and run \(i\) used random seed \(28+i\) for parameter initialization and other
stochastic operations.
The four non-test folds were pooled and divided into training and validation
subsets using an 80:20 split. Scaling parameters were calculated from the
training and validation subsets and then held fixed when processing the
corresponding test subset.

Table~\ref{tab:training_hyperparameters} lists the complete hyperparameter choices used for
the final models. Learning-rate reduction and early stopping were both based
on validation loss. Weighted sampling was used only while constructing
training batches; validation and test samples retained their natural class
distributions. The five randomly initialized encoder baselines used the same
per-run seed convention but were not trained.
\begin{table}[t]
\centering
\small
\setlength{\tabcolsep}{3pt}
\begin{tabularx}{\columnwidth}{
@{}
l
>{\raggedright\arraybackslash}X
>{\raggedright\arraybackslash}X
@{}
}
\toprule
\textbf{Setting} & \textbf{ResNet14} & \textbf{PointNet-style} \\
\midrule
Loss
& Cross-entropy
& Cross-entropy \\

Optimizer
& AdamW
& AdamW \\

Initial learning rate
& \(5\times10^{-4}\)
& \(5\times10^{-4}\) (\(^{16}\mathrm{O}+\alpha\));
  \(5\times10^{-6}\) (\(^{20}\mathrm{Mg}\)) \\

Weight decay
& \(10^{-4}\)
& \(10^{-4}\) \\

Batch size
& 128
& 128 \\

Learning-rate scheduler
& Reduce on plateau
& Reduce on plateau \\

Scheduler parameters
& Factor 0.1; patience 10; relative threshold \(10^{-3}\)
& Factor 0.1; patience 10; relative threshold \(10^{-3}\) \\

Minimum learning rate
& \(5\times10^{-8}\)
& \(5\times10^{-8}\) \\

Maximum epochs
& 200
& 200 \\

Early-stopping patience
& 30 epochs
& 30 epochs \\

Class imbalance
& Weighted training sampler
& Weighted training sampler \\

Gradient clipping
& Maximum norm 1.0
& None \\

Coordinate translation
& Training only; \(\mathcal{U}(-0.9,0.9)\) per dimension
& None \\
\bottomrule
\end{tabularx}
\caption{Optimization, regularization, and augmentation settings used to
train the ResNet14 and PointNet-style encoders. Settings were used across
encoder-training datasets unless otherwise indicated.}
\label{tab:training_hyperparameters}
\end{table}
\end{document}